%% file: encode_rule_into_knowledge_graph_embedding.tex
\documentclass[sigconf]{acmart}

\usepackage{booktabs} 
\usepackage{epstopdf}
\usepackage{latexsym}
\usepackage{url}
\usepackage{amsmath}
\usepackage{amssymb}
\usepackage{graphicx}
\usepackage{caption}
\usepackage{subfigure}
\usepackage{multirow}
\usepackage{enumitem}
\usepackage{bm}
\usepackage{changepage}
\usepackage{algorithm}
\usepackage{algorithmic}
\usepackage{array}
\usepackage{natbib}

\begin{document}
\title{Logic Rules Powered Knowledge Graph Embedding}

\author{Pengwei Wang}
\affiliation{%
  \institution{Alibaba Group}
}
\email{hoverwang.wpw@alibaba-inc.com}

\author{Dejing Dou}
\affiliation{%
  \institution{University of Oregon}
}
\email{dou@cs.uoregon.edu}

\author{Fangzhao Wu}
\affiliation{%
  \institution{Microsoft}
}
\email{fangzwu@microsoft.com}

\author{Nisansa de Silva}
\affiliation{%
  \institution{University of Oregon}
}
\email{nisansa@cs.uoregon.edu}

\author{Lianwen Jin}
\affiliation{%
  \institution{South China University of Technology}
}
\email{lianwen.jin@gmail.com}

\begin{abstract}
Large scale knowledge graph embedding has attracted much attention from both academia and industry in the field of Artificial Intelligence. However, most existing methods concentrate solely on fact triples contained in the given knowledge graph. Inspired by the fact that logic rules can provide a flexible and declarative language for expressing rich background knowledge, it is natural to integrate logic rules into knowledge graph embedding, to transfer human knowledge to entity and relation embedding, and strengthen the learning process. In this paper, we propose a novel logic rule-enhanced method which can be easily integrated with any translation based knowledge graph embedding model, such as TransE \cite{Bordes:2013}. We first introduce a method to automatically mine the logic rules and corresponding confidences from the triples. And then, to put both triples and mined logic rules within the same semantic space, all triples in the knowledge graph are represented as first-order logic. Finally, we define several operations on the first-order logic and minimize a global loss over both of the mined logic rules and the transformed first-order logics. We conduct extensive experiments for link prediction and triple classification on three datasets: WN18, FB166, and FB15K. Experiments show that the rule-enhanced method can significantly improve the performance of several baselines. The highlight of our model is that the filtered Hits@1, which is a pivotal evaluation in the knowledge inference task, has a significant improvement (up to 700\% improvement).

\end{abstract}

%
%
%


\maketitle

\input{samplebody-conf}

\bibliographystyle{ACM-Reference-Format}
\bibliography{sigproc}

\end{document}

%% file: samplebody-conf.tex
\section{Introduction}

Knowledge graphs such as Freebase\footnote{https://developers.google.com/freebase/} \cite{Bollacker:1}, Wordnet\footnote{https://wordnet.princeton.edu/} \cite{Miller:1} and YAGO\footnote{http://www.mpi-inf.mpg.de/departments/databases-and-information-systems/research/yago-naga/yago/\#c10444} \cite{Suchanek:1} play a pivotal role in many Artificial Intelligence related tasks, such as web search, automatic question answering systems, etc. A knowledge graph is a multi-relational data composed of entities as nodes and relations as different types of edges. A fact triple is stored in the form of (head entity, relation, tail entity) (denoted as ($h, r, t$)), e.g., {\fontfamily{qcr}\selectfont (Washington, isCapitalof, USA)}. Although a significant number of large scale knowledge graphs have been constructed, the symbolic nature of such triples makes knowledge graph hard to manipulate, especially in knowledge inference tasks \cite{Guo:emnlp2016}.

Recently, a new approach, namely knowledge graph embedding, has been proposed to handle the symbolic nature problems. The goal of the knowledge graph embedding is to embed a knowledge graph into a continuous vector space while preserving certain properties of the original graph \cite{Socher:2013,Bordes:2013,Weston:2013,Bordes:2011,Chang:2014}. For example, the entity $h$ (or $t$) is represented as a point $\mathbf{h}$ (or $\mathbf{t}$) in vector space, and the relation $r$ is represented as a translation operation $\mathbf{r}$ in the same vector space. The embedding representations contain rich semantic information of entities and relations, and global knowledge graph information.

However, most existing methods \cite{Bordes:2013,Wang:2014a} concentrate solely on fact triples contained in the given knowledge graph but not on other background knowledge. Logic rules provide a flexible and declarative language for expressing structured and rich background knowledge \cite{Hu:1}. It is therefore desirable to integrate logic rules into knowledge graph embedding, to transfer human knowledge to entity and relation embedding, and strengthen the learning process. Recently, \citet{Wang:2015} and \citet{Wei:2015} attempted to leverage the rules into the knowledge graph embedding. However, the rules are used in the post-processing step, which is separate from the knowledge graph embedding process. \citet{Guo:emnlp2016} first proposed jointly embedding rules and triples. However, they manually selected the rules in the rule mining process, and they did not represent rules and triples in the same space. Although these models are effective in the experiments, there are three limitations in most of the models.

\begin{enumerate}
  \item The logic rules are underused in most knowledge graph embedding models.
  \item Rules need to be manually selected, which leads to the fact that it is hard to extend these methods to a large scale knowledge graph.
  \item Rule are encoded in the form of true value, which leads to the one-to-many mapping phenomenon. For example, one encoding form can map several rules including correct and incorrect rules.
  \item Algebraic operations of logic symbols are inconsistent in triples and rules. Thus, it is hard to efficiently jointly embed the rules and the knowledge graph.
\end{enumerate}

In this paper, we propose a novel logic rule-enhanced method which can be easily integrated with any translation based knowledge graph embedding model, such as TransE \cite{Bordes:2013}. Firstly, we introduce a method to automatically mine the logic rules and the corresponding confidences from the triples. In general, rules are composed by several components. In this paper, we consider three types of rules: inference rules, transitivity rules, and antisymmetry rules. Inference rules denote that one relation can imply another relation where the former relation could be a subproperty of the latter relation. Transitivity rules denote the combination of two relations can imply the third relation. Antisymmetry rules denote one relation can imply another relation which has opposite meaning. Only rules with confidences greater than a threshold are used in our rule-enhanced knowledge graph embedding. To make the algebraic operations consistency between triples and rules, we propose to map all of the triples and rules into first-order logics. For example, a triple ($h, r, t$) can be represented as $r(h)\Rightarrow t$. To guarantee the encoding form of a rule and the rule have one-to-one mapping relation, we propose a general interaction operation for rules to make the components within rules directly interact in the vector space. Finally, we define operations for all of the logical symbols used in the transformed first-order logic. A global loss over the transformed first-order logics is minimized in the knowledge graph embedding process.


We evaluate our rule-enhanced method on link prediction and triple classification on three datasets: WN18, FB166, and FB15K. Experiments show that our proposed rule-enhanced method can significantly improve the performance of several baseline models.  Particularly, the filtered Hits@1, which is a pivotal evaluation in the knowledge inference task, has a great improvement (up to 700\% improvement). To summarize, the contributions of this paper are as follows:
\begin{enumerate}
  \item We propose a novel logic rule-enhanced method which can be easily integrated with any translation based knowledge graph embedding model.
  \item We introduce a method to automatically mine logic rules and corresponding confidences from the triples in a given knowledge graph.
  \item We propose to transform the triples and logic rules into the same first-order logic space and encode the rules in the vector space.
  \item Experiments show that our method brings a great improvement in the filtered Hits@1 evaluation.
\end{enumerate}


\section{Problem Definition}
In this section, we first give a formal definition of the knowledge graph embedding. A knowledge graph $\mathcal{G}$ is defined as a set of triples $\mathcal{K}={(h,r,t)}$, with each triple composed of two entities $h,t\in\mathcal{E}$ and their corresponding relation $r\in\mathcal{R}$. $\mathcal{E}$ denotes the entity vocabulary in $\mathcal{G}$ and $\mathcal{R}$ denotes the relation vocabulary in $\mathcal{G}$. $h$ denotes the head entity of a triple and $t$ denotes the tail entity of a triple. The goal of knowledge graph embedding is to learn the embedding representations of entities and relations for easily computing in continuous space, e.g., computing the similarity of two entities. We use bold letters $\mathbf{h},\mathbf{r},\mathbf{t}$ to denote the embedding of $h,r,t$, where $\mathbf{h},\mathbf{r},\mathbf{t}\in\mathbb{R}^d$.

\section{Rule Extraction}
We introduce three types of rules, which are used in our proposed model.
\begin{itemize}
  \item \textbf{Rule 1 (inference rule).} An inference rule is in the form of $\forall h,t : (h,r_1,t)\Rightarrow (h,r_2,t)$. The relation $r_1$ can imply relation $r_2$, which denotes that any two entities linked by $r_1$ should also be linked by $r_2$. For example, {\fontfamily{qcr}\selectfont (Washington, isCapitalof, USA)} $\Rightarrow$ {\fontfamily{qcr}\selectfont(Washington, isLocatedin, USA)}. An inference rule is directed. We define the relation $r_2$ to be the concept of the relation $r_1$ ($r_1$ is an instance of $r_2$).
  \item \textbf{Rule 2 (transitivity rule).} A transitivity rule is in the form of $\forall e_1,e_2,e_3 : [(e_1,r_1,e_2)+(e_2,r_2,e_3)\Rightarrow(e_1,r_3,e_3)]$. The orderly conjunction of relation $r_1$ and relation $r_2$ can imply relation $r_3$. The transitivity rule denotes that if $e_1$ and $e_2$ are linked by relation $r_1$ and $e_2$ and $e_3$ are linked by relation $r_2$, $e_1$ and $e_3$ will be linked by relation $r_3$. For example, {\fontfamily{qcr}\selectfont (George W. Bush, Nationality, USA) } + {\fontfamily{qcr}\selectfont (USA, Official\_Language, English)} $\Rightarrow$ {\fontfamily{qcr}\selectfont (George W. Bush, Language, English)}.
  \item \textbf{Rule 3 (antisymmetry rule).} An antisymmetry rule is in the form of $\forall h,t : (h,r_1,t)\Leftrightarrow (t,r_2,h)$. The relation $r_1$ can imply the antisymmetry relation $r_2$, which denotes that two relations $r_1$ and $r_2$ are antisymmetrical. For example, {\fontfamily{qcr}\selectfont (apple, hypernym, fruit)} $\Leftrightarrow$ {\fontfamily{qcr}\selectfont(fruit, hyponym, apple)}. An antisymmetry rule is undirected.
\end{itemize}

\begin{figure*}[t]
  \centering
  \includegraphics[width=16.3cm]{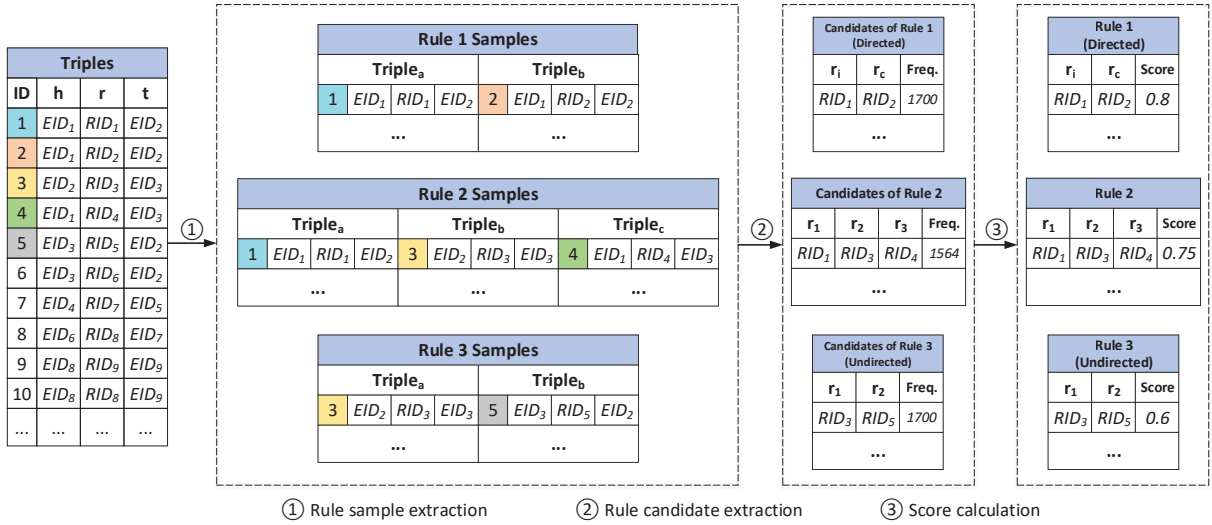}\\
  \caption{The flowchart of rule extraction.}\label{fig:rule_extraction}
\end{figure*}

As shown in Figure \ref{fig:rule_extraction}, the input of this framework is triples of the knowledge graph $\mathcal{G}$, and the output is the ground rules of the three types with corresponding scores. There are three steps in this framework.
\subsection{Rule Sample Extraction}
In this step, we extract the rule samples from given triples. Here, we define the rule samples as the triple combinations that meet the conditions of rules. For example, a rule 1 sample composed the ID 1 triple and ID 2 triple in the "Rule 1 Samples" table of Figure 1 can meet the inference rule because the two triples have the same head entity and tail entity; a rule 2 sample composed of the ID 1 triple, ID 3 triple and ID 4 triple in the "Rule 2 Samples" table of Figure 1 can meet the transitivity rule because of the sequential entities (\emph{EID}$_1$, \emph{EID}$_2$, and \emph{EID}$_3$); a rule 3 sample composed of the ID 3 triple and ID 5 triple in the "Rule 3 Samples" table of Figure 1 can meet the antisymmetry rule because the two triples have the antisymmetric head entity and tail entity. Many rule samples belonging to the above three types of rules are mined using this manner.

\subsection{Rule Candidate Extraction}
We extract some candidates for rules and do some statistics according to the rule samples. For example, (\emph{RID}$_1$ and \emph{RID}$_2$) in the "Candidates of Rule 1" table of Figure 1 are extracted from each sample of rule 1; (\emph{RID}$_1$, \emph{RID}$_3$, and \emph{RID}$_4$) in the "Candidates of Rule 2" table of Figure 1 are extracted from each sample of rule 2; (\emph{RID}$_3$ and \emph{RID}$_5$) in the "Candidates of Rule 3" table of Figure 1 are extracted from each sample of rule 3. All candidates of three types of rules are extracted from the rule samples using this way. Then, we rank the candidates in descending order according to the frequencies. Because the inference rule is directed, we need to distinguish the concept relation and the instance relation. The relations in Freebase are hierarchical. For example, the {\fontfamily{qcr}\selectfont /location/country/capital} relation has three diminishing levels ({\fontfamily{qcr}\selectfont location, country, and capital}). To judge the concept relation between {\fontfamily{qcr}\selectfont /location/country/capital} and {\fontfamily{qcr}\selectfont /location/location/contains}, we only judge the concept from the second level, namely {\fontfamily{qcr}\selectfont country} and {\fontfamily{qcr}\selectfont location}, since the first levels are the same. In Freebase, each entity has several object types, as shown in Table \ref{tab:type}. The object type is also hierarchical. We can get the concept level from the object type. For example, we obtain that {\fontfamily{qcr}\selectfont location} is the concept of {\fontfamily{qcr}\selectfont country} from the object type {\fontfamily{qcr}\selectfont location.country}. Thus, the relation {\fontfamily{qcr}\selectfont /location/location/contains} is the concept of the relation {\fontfamily{qcr}\selectfont /location/country/capital}. In addition, we can also get the concept of an entity by leveraging Probase \footnote{https://concept.msra.cn/Home/Download} \cite{Wu:1}. Given a word, Probase will provide the concepts associated with the word. More usages of Probase can be found in its official website.

\begin{table}[t]
\begin{minipage}{1\linewidth}
\centering
\renewcommand{\multirowsetup}{\centering}
\renewcommand\arraystretch{1}
\caption{Entity object types.} \label{tab:type}
\begin{tabular}{lc}
\toprule
\textbf{Relation}&\textbf{Concept-Instance Pairs}\\
\midrule
location.country&(location-country)\\
people.profession&(people-profession)\\
music.songwriter&(music-songwriter)\\
sports.boxer&(sports-boxer)\\
book.magazine&(book-magazine)\\
\bottomrule
\end{tabular}
\vspace{-15pt}
\end{minipage}
\end{table}

\subsection{Score Calculation}

Many candidate rules are not correct. Not all results inferred by the candidate rules are reasonable. In KALE \cite{Guo:emnlp2016}, the wrong rules ranking at the top are manually filtered. It is hard to leverage KALE in a large scale knowledge graph, which has a large number of relations. We propose a novel method to automatically select the rules from the candidate pool. Algorithm \ref{alg:score} summarizes the proposed method. We first generate all of the new triples according to each candidate rule using the function \emph{GetNewtriples}(), which is introduced as follows.

\begin{itemize}
    \item Inference candidate rule: Given a candidate rule $r_1 \Rightarrow r_2$, if a triple has the relation $r_1$, such as ($h$,$r_1$,$t$), a new triple ($h$,$r_2$,$t$) can be generated. Each candidate can generate triples in the same manner.
    \item Transitivity candidate rule: Given a candidate rule $(r_1\!+\!r_2)$$\Rightarrow$$r_3$, if there are two triples ($e_1$,$r_1$,$e_2$) and ($e_2$,$r_2$,$e_3$), a new triple ($e_1$,$r_3$,$e_3$) can be generated. Each candidate can generate triples in the same manner.
    \item Antisymmetry candidate rule: Given a candidate rule $r_1 \Leftrightarrow r_2$, if a triple has the relation $r_1$, such as ($h$,$r_1$,$t$), a new triple ($t$,$r_2$,$h$) can be generated. Each candidate can generate triples in the same manner.
\end{itemize}

As shown in Algorithm 1, we can calculate the percentage $\alpha_i$ about the new triples that exist in original triples $\mathcal{K}$ for each candidate rule. We only select the rules whose percentages are greater than the threshold $\tau$. Then, we generate the ground rules for each rule in the triples $\mathcal{K}$. For example, given a rule $r_1$$\Rightarrow$$r_2$, we can generate a corresponding ground rule $(h_m,r_1,t_n)$$\Rightarrow$$(h_m,r_2,t_n)$ from a triple $(h_m,r_1,t_n)$. Finally, we obtain all of the ground rules for each rule.

\renewcommand{\algorithmicrequire}{\textbf{Input:}}
\renewcommand{\algorithmicensure}{\textbf{Output:}}
\algsetup{indent=2em}
\begin{algorithm}[t]
\caption{Score Calculation}
\label{alg:score}
\small
\begin{algorithmic}[1]
    \REQUIRE Candidate rules $R_a$, triples $\mathcal{K}$, threshold $\tau$
    \ENSURE Ultimate rules $R_u$

\STATE $T$ $\gets$ $\emptyset$
\FOR {each rule $r$ in $R_a$}
\STATE $T_n$ $\gets$ GetNewtriples($r$)
\STATE $T$ $\gets$ $T\cup T_n$
\ENDFOR

\STATE $A$ $\gets$ $\emptyset$
\FOR {each new triple set $T_n$ in $T$}
\STATE $\alpha_i=\frac{\#(T_n\cup \mathcal{K})}{\#{T_n}}$
\STATE $A$ $\gets$ $A\cup \alpha_i$
\ENDFOR

\FOR {$i=1$ \TO $\#(R_a)$}
\IF {$A[i]\geq \tau $}
\STATE $R_u$ $\gets$ $R_u\cup R_a[i]$
\ENDIF
\ENDFOR
\end{algorithmic}
\end{algorithm}

\section{Rule-Enhanced Knowledge Graph Embedding}
The proposed rule-enhanced knowledge graph embedding method can be easily used in any translation based knowledge graph embedding model, such as TransE \cite{Bordes:2013}. In this paper, we apply it into not only TransE \cite{Bordes:2013}, but also TransH \cite{Wang:2014a} and TransR \cite{Lin:2015a}.

\subsection{Rule-Enhanced TransE Model}
We follow TransE \cite{Bordes:2013} to model the triples. TransE represents a relation by a translation vector $\mathbf{r}$ so that the pair of the two embedded entities in a triple $(h, r, t)$ can be linked by $\mathbf{r}$. The score function $s_1(h, r, t)$ of TransE is:
\begin{equation}\label{eq:transe}
s_1(h, r, t)=\parallel\mathbf{h}\!+\!\mathbf{r}\!-\!\mathbf{t}\parallel_{l_{1/2}},
\end{equation}
where $\mathbf{h}$ and $\mathbf{t}$ are the learned entity embeddings. $\mathbf{r}$ is the learned relation embedding, and $l_{1/2}$ denotes the $L_1$-norm or $L_2$-norm respectively.

Then, we try to model the ground rules into the knowledge graph embedding process. As shown in Table \ref{tab:fol}, we formulate the knowledge graph and the three types of ground rules as first-order logic. In the reference ground rule (the second row in Table \ref{tab:fol}), $C$ denotes the concept of entity $h$ in the triple $(h,r_2,t)$, which can be obtained in Section 3. The head entity $h$ in two relations $r_1$ and $r_2$ has different concepts. For example, in this inference ground rule {\fontfamily{qcr}\selectfont (Washington, isCapitalof, USA)} $\Rightarrow$ {\fontfamily{qcr}\selectfont(Washington, isLocatedin, USA)}, the concepts of the first and second {\fontfamily{qcr}\selectfont Washington} are capital and location respectively. In a inference ground rule, we assume the head entity must be an instance of the concept of the head in second relation if the first relation $r_1$ want to deduce the second relation $r_2$. In addition, we assume that the entity $h$ can deduce the entity $t$ after that the entity $h$ is translated by relation $r$.

\begin{table*}[t]
\begin{minipage}{1\linewidth}
\centering
\renewcommand{\multirowsetup}{\centering}
\renewcommand\arraystretch{1}
\caption{The form of first-order logic.} \label{tab:fol}
\begin{tabular}{ll}
\toprule
\textbf{Triple and ground rule}&\textbf{The format of first-order logic.}\\
\midrule
$(h, r, t)$&$r(h)\Rightarrow t$\\
$(h, r_1, t)\Rightarrow(h, r_2, t)$&$[(h\in C) \wedge [r_1(h) \Rightarrow t]] \Rightarrow [r_2(h) \Rightarrow t]$\\
$(e_1, r_1, e_2)+(e_2, r_2, e_3)\Rightarrow(e_1, r_3, e_3)$&$[[r_1(e_1) \Rightarrow e_2] \wedge [r_2(e_2) \Rightarrow e_3] ] \Rightarrow [r_3(e_1) \Rightarrow e_3]$\\
$(h, r_1, t)\Leftrightarrow(t, r_2, h)$&$[[ r_1(h) \Rightarrow t] \Rightarrow [r_2(t) \Rightarrow h]]\wedge[[ r_2(t) \Rightarrow h] \Rightarrow [r_1(h) \Rightarrow t]]$\\
\bottomrule
\end{tabular}
\end{minipage}
\end{table*}

After formulating the knowledge graph and the three types of ground rules as first-order logic, some logical symbols, such as $r(\cdot)$, $\Rightarrow$, $\in$, and $\wedge$,  are defined in Table \ref{tab:exp}. The $\otimes$ denotes pairwise multiplication. The $a$ and $b$ can be an entity, a triple or a combination of triples.

\begin{table}[h]
\begin{minipage}{1\linewidth}
\centering
\renewcommand{\multirowsetup}{\centering}
\renewcommand\arraystretch{1}
\caption{Mathematical expression of first-order logic.} \label{tab:exp}
\begin{tabular}{ll}
\toprule
\textbf{First-order logic}&\textbf{Mathematical expression}\\
\midrule
$r(h)$&$ \mathbf{r}+\mathbf{h} $\\
$ a \Rightarrow b $&$ \mathbf{a}-\mathbf{b} $\\
$ h \in C $&$ \mathbf{h}\cdot \mathbf{C} $ ($\mathbf{C}$ is a matrix)\\
$ a \wedge b $&$ \mathbf{a} \otimes \mathbf{b} $\\
$ a \Leftrightarrow b $&$ (\mathbf{a}-\mathbf{b})\otimes(\mathbf{a}-\mathbf{b}) $\\
\bottomrule
\end{tabular}
\end{minipage}
\end{table}

We considers three types of ground rules. The first type is $\forall h,t$$:$ $(h,r_1,t)$$\Rightarrow$$(h,r_2,t)$. Given a ground rule $f\triangleq(h_m,r_1,t_n)$ $\Rightarrow$ $(h_m,r_2,t_n)$, the score function is calculated by:
\begin{equation}
\begin{split}
s_2(f)=\parallel\!(\mathbf{h}_m\!\cdot\!\mathbf{C})\!\otimes\!(\mathbf{h}_m\!+\!\mathbf{r}_1\!-\!\mathbf{t}_n)\!-\!(\mathbf{h}_m\!+\!\mathbf{r}_2\!-\!\mathbf{t}_n)\!\parallel_{l_{1/2}}.
\end{split}
\end{equation}

The second type is $\forall e_1,e_2,e_3 : [(e_1,r_1,e_2)+(e_2,r_2,e_3)\Rightarrow(e_1,r_3,e_3)]$. Given a ground rule $f\triangleq[(e_l,r_1,e_m)+(e_m,r_2,e_n)\Rightarrow(e_l,r_3,e_n)]$, the score function is calculated by:
\begin{equation}
\begin{split}
s_3(f)=&\parallel[(\mathbf{e}_l+\mathbf{r}_1-\mathbf{e}_m)\otimes(\mathbf{e}_m+\mathbf{r}_2-\mathbf{e}_n)]\\
&-(\mathbf{e}_l+\mathbf{r}_3-\mathbf{e}_n)\parallel_{l_{1/2}}.
\end{split}
\end{equation}

The third type is $\forall h,t : (h,r_1,t)\Rightarrow (t,r_2,h)$. Given a ground rule $f\triangleq(h_m,r_1,t_n)\Rightarrow(t_n,r_2,h_m)$, the score function is calculated by:
\begin{equation}
\begin{aligned}
s_4(f)=\parallel(\mathbf{TR}_f-\mathbf{TR}_b)\otimes(\mathbf{TR}_b-\mathbf{TR}_f)\parallel_{l_{1/2}},
\end{aligned}
\end{equation}
\begin{equation}
\mathbf{TR}_f=\mathbf{h}_m+\mathbf{r}_1-\mathbf{t}_n,\quad\mathbf{TR}_b=\mathbf{t}_n+\mathbf{r}_2-\mathbf{h}_m.
\end{equation}

\subsection{Rule-Enhanced TransH Model}

To address the issue of TransE when modeling N-to-1, 1-to-N and N-to-N relations, \citet{Wang:2014a} proposed TransH to enable an entity to have distinct embeddings when involved in different relations. For a relation $r$, TransH models the relations as a vector $\mathbf{r}$ on a hyperplane with $\mathbf{w}_r$ as the normal vector. The score function $s_1(h,r,t)$ of TransH is:
\begin{equation}\label{eq:transh}
s_1(h, r, t)=\parallel(\mathbf{h}-\mathbf{w}_r^{\top}\mathbf{h}\mathbf{w}_r)\!+\!\mathbf{r}\!-\!(\mathbf{t}-\mathbf{w}_r^{\top}\mathbf{t}\mathbf{w}_r)\parallel_{l_{1/2}},
\end{equation}
where $\mathbf{h}$,$\mathbf{t}$ are the learned entity embeddings, $\mathbf{r}$ is the learned relation embedding and $\mathbf{w}_r$ is the learned normal vector.

As introduced in section 4.1, we apply the three types of ground rules in the TransH model. For the inference ground rule, given a ground rule $f\triangleq(h_m,r_1,t_n)\Rightarrow(h_m,r_2,t_n)$, the score function is calculated by:
\begin{equation}
s_2(f)=\parallel\!(\mathbf{h}_m\!\cdot \mathbf{C})\otimes(\mathbf{h}^m_{\bot_1}+\mathbf{r}_1\!-\!\mathbf{t}^n_{\bot_1})-(\mathbf{h}^m_{\bot_2}+\mathbf{r}_2-\mathbf{t}^n_{\bot_2})\!\parallel_{l_{1/2}},
\end{equation}
\begin{equation}
\mathbf{h}^m_{\bot_1}=\mathbf{h}_m-\mathbf{w}^{\top}_{r_1}\mathbf{h}_m\mathbf{w}_{r_1}, \quad \mathbf{t}^n_{\bot_1}=\mathbf{t}_n-\mathbf{w}^{\top}_{r_1}\mathbf{t}_n\mathbf{w}_{r_1},
\end{equation}
\begin{equation}
\mathbf{h}^m_{\bot_2}=\mathbf{h}_m-\mathbf{w}^{\top}_{r_2}\mathbf{h}_m\mathbf{w}_{r_2}, \quad \mathbf{t}^n_{\bot_2}=\mathbf{t}_n-\mathbf{w}^{\top}_{r_2}\mathbf{t}_n\mathbf{w}_{r_2}.
\end{equation}

Given a transitivity ground rule $f\triangleq[(e_l,r_1,e_m)\!+\!(e_m,r_2,e_n)$ $\Rightarrow$$(e_l,r_3,e_n)]$, the score function is calculated by:
\begin{equation}
\begin{split}
s_3(f)=&\parallel[(\mathbf{e}^l_{\bot_1}+\mathbf{r_1}-\mathbf{e}^m_{\bot_1})\otimes(\mathbf{e}^m_{\bot_2}+\mathbf{r_2}-\mathbf{e}^n_{\bot_2})]\\
&-(\mathbf{e}^l_{\bot_3}+\mathbf{r_3}-\mathbf{e}^n_{\bot_3})\parallel_{l_{1/2}},
\end{split}
\end{equation}
\begin{equation}
\mathbf{e}^l_{\bot_1}=\mathbf{e}_l-\mathbf{w}^{\top}_{r_1}\mathbf{e}_l\mathbf{w}_{r_1}, \quad \mathbf{e}^m_{\bot_1}=\mathbf{e}_m-\mathbf{w}^{\top}_{r_1}\mathbf{e}_m\mathbf{w}_{r_1},
\end{equation}
\begin{equation}
\mathbf{e}^m_{\bot_2}=\mathbf{e}_m-\mathbf{w}^{\top}_{r_2}\mathbf{e}_m\mathbf{w}_{r_2}, \quad \mathbf{e}^n_{\bot_2}=\mathbf{e}_n-\mathbf{w}^{\top}_{r_2}\mathbf{e}_n\mathbf{w}_{r_2},
\end{equation}
\begin{equation}
\mathbf{e}^l_{\bot_3}=\mathbf{e}_l-\mathbf{w}^{\top}_{r_3}\mathbf{e}_l\mathbf{w}_{r_3}, \quad \mathbf{e}^n_{\bot_3}=\mathbf{e}_n-\mathbf{w}^{\top}_{r_3}\mathbf{e}_n\mathbf{w}_{r_3}.
\end{equation}

Given a antisymmetry ground rule $f\triangleq(h_m,r_1,t_n)\Rightarrow(t_n,r_2,h_m)$, the score function is calculated by:
\begin{equation}
\begin{aligned}
s_4(f)=\parallel(\mathbf{TR}_f-\mathbf{TR}_b)\otimes(\mathbf{TR}_b-\mathbf{TR}_f)\parallel_{l_{1/2}},
\end{aligned}
\end{equation}
\begin{equation}
\mathbf{TR}_f=\mathbf{h}^m_{\bot_1}+\mathbf{r_1}-\mathbf{t}^n_{\bot_1},\quad\mathbf{TR}_b=\mathbf{t}^n_{\bot_2}+\mathbf{r_2}-\mathbf{h}^m_{\bot_2}.
\end{equation}
\begin{equation}
\mathbf{h}^m_{\bot_1}=\mathbf{h}_m-\mathbf{w}^{\top}_{r_1}\mathbf{h}_m\mathbf{w}_{r_1}, \quad \mathbf{t}^n_{\bot_1}=\mathbf{t}_n-\mathbf{w}^{\top}_{r_1}\mathbf{t}_n\mathbf{w}_{r_1},
\end{equation}
\begin{equation}
\mathbf{h}^m_{\bot_2}=\mathbf{h}_m-\mathbf{w}^{\top}_{r_2}\mathbf{h}_m\mathbf{w}_{r_2}, \quad \mathbf{t}^n_{\bot_2}=\mathbf{t}_n-\mathbf{w}^{\top}_{r_2}\mathbf{t}_n\mathbf{w}_{r_2}.
\end{equation}

\subsection{Rule-Enhanced TransR Model}
The entities in TransR \cite{Lin:2015a} are mapped into vectors in different relation space embedding according to a relation. Thus, the score function $s_1(h,r,t)$ of TransR is:
\begin{equation}\label{eq:transr}
s_1(h, r, t)=\parallel\mathbf{h}\mathbf{M}_r+\mathbf{r}-\mathbf{t}\mathbf{M}_r\parallel_{l_{1/2}},
\end{equation}
where $\mathbf{h}$,$\mathbf{t}$ are the learned entity embeddings, $\mathbf{r}$ is the learned relation embedding and $\mathbf{M}_r$ is the learned projection matrix.

As introduced in section 4.1, we apply the three types of ground rules in the TransH model. For the inference ground rule, given a ground rule $f\triangleq(h_m,r_1,t_n)\Rightarrow(h_m,r_2,t_n)$, the score function is calculated by:
\begin{equation}
s_2(f)=\parallel(\mathbf{h}_m\!\cdot \mathbf{C})\otimes(\mathbf{h}^m_{r_1}+\mathbf{r}_1-\mathbf{t}^n_{r_1})\!-\!(\mathbf{h}^m_{r_2}+\mathbf{r}_2-\mathbf{t}^n_{r_2})\parallel_{l_{1/2}},
\end{equation}
\begin{equation}
\mathbf{h}^m_{r_1}=\mathbf{h}_m\mathbf{M}_{r_1}, \qquad \mathbf{t}^n_{r_1}=\mathbf{t}_n\mathbf{M}_{r_1},
\end{equation}
\begin{equation}
\mathbf{h}^m_{r_2}=\mathbf{h}_m\mathbf{M}_{r_2}, \qquad \mathbf{t}^n_{r_2}=\mathbf{t}_n\mathbf{M}_{r_2}.
\end{equation}

Given a transitivity ground rule $f$$\triangleq$$[(e_l,r_1,e_m)\!+\!(e_m,r_2,e_n)$ $\Rightarrow(e_l,r_3,e_n)]$, the score function is calculated by:
\begin{equation}
\begin{split}
s_3(f)=&\parallel[(\mathbf{e}^l_{r_1}+\mathbf{r_1}-\mathbf{e}^m_{r_1})\otimes(\mathbf{e}^m_{r_2}+\mathbf{r_2}-\mathbf{e}^n_{r_2})]\\
&-(\mathbf{e}^l_{r_3}+\mathbf{r_3}-\mathbf{e}^n_{r_3})\parallel_{l_{1/2}},
\end{split}
\end{equation}
\begin{equation}
\mathbf{e}^l_{r_1}=\mathbf{e}_l\mathbf{M}_{r_1}, \qquad \mathbf{e}^m_{r_1}=\mathbf{e}_m\mathbf{M}_{r_1},
\end{equation}
\begin{equation}
\mathbf{e}^m_{r_2}=\mathbf{e}_m\mathbf{M}_{r_2}, \qquad \mathbf{e}^n_{r_2}=\mathbf{e}_n\mathbf{M}_{r_2},
\end{equation}
\begin{equation}
\mathbf{e}^l_{r_3}=\mathbf{e}_l\mathbf{M}_{r_3}, \qquad \mathbf{e}^n_{r_3}=\mathbf{e}_n\mathbf{M}_{r_3}.
\end{equation}

Given a antisymmetry ground rule $f\triangleq(h_m,r_1,t_n)\Rightarrow(t_n,r_2,h_m)$, the score function is calculated by:
\begin{equation}
\begin{aligned}
s_4(f)=\parallel(\mathbf{TR}_f-\mathbf{TR}_b)\otimes(\mathbf{TR}_b-\mathbf{TR}_f)\parallel_{l_{1/2}},
\end{aligned}
\end{equation}
\begin{equation}
\mathbf{TR}_f=\mathbf{h}^m_{r_1}+\mathbf{r_1}-\mathbf{t}^n_{r_1},\quad\mathbf{TR}_b=\mathbf{t}^n_{r_2}+\mathbf{r_2}-\mathbf{h}^m_{r_2}.
\end{equation}
\begin{equation}
\mathbf{h}^m_{r_1}=\mathbf{h}_m\mathbf{M}_{r_1}, \quad \mathbf{t}^n_{r_1}=\mathbf{t}_n\mathbf{M}_{r_1},
\end{equation}
\begin{equation}
\mathbf{h}^m_{r_2}=\mathbf{h}_m\mathbf{M}_{r_2}, \quad \mathbf{t}^n_{r_2}=\mathbf{t}_n\mathbf{M}_{r_2}.
\end{equation}

\section{Global Objective Function}

We minimize a global loss over triples and ground rules to learn the entity and relation embedding representation. Our training dataset $\mathcal{F}$ contains the original triples and the three types of ground rules. Then, we minimize the following loss function to learn the entity embedding and relation embedding from ground rules and original triples:
\begin{equation}\label{eq:loss}
\begin{split}
\min_{\mathbf{e},\mathbf{r}}\sum_{f^+\in\mathcal{F}}\sum_{f^-\in\mathcal{N}_{f^+}}[S]_+,
\end{split}
\end{equation}
\begin{equation}
\begin{split}
S=I_1(f^+)\cdot(\gamma+s_1(f^+)-s_1(f^-))\\
+I_2(f^+)\cdot(\gamma+s_2(f^+)-s_2(f^-))\\
+I_3(f^+)\cdot(\gamma+s_3(f^+)-s_3(f^-))\\
+I_4(f^+)\cdot(\gamma+s_4(f^+)-s_4(f^-)),
\end{split}
\end{equation}
where $s_1()$, $s_2()$, $s_3()$, and $s_4()$ are introduced in Section 4. $I_n(f)$ denotes the gate, which is shown in Table \ref{tab:gate}. When a sample is selected, only one gate is active each time. $f^+$ is a positive sample, such as a triple or a ground rule, $f^-$ is a negative sample constructed by corrupting $f^+$, $\gamma>0$ is a margin separating positive and negative samples, and $[x]_+\triangleq\max\{0,x\}$ denotes the positive part of $x$. The following constrains are considered when we minimize the loss function:
\begin{equation}
\forall e\in \mathcal{E}, ||\mathbf{e}||_2\leq 1, \qquad \forall r\in \mathcal{R}, ||\mathbf{r}||_2\leq 1.
\end{equation}


\begin{table}[t]
\begin{minipage}{1\linewidth}
\centering
\renewcommand{\multirowsetup}{\centering}
\renewcommand\arraystretch{1}
\caption{Gate state.} \label{tab:gate}
\begin{tabular}{ll}
\toprule
\textbf{Sample}&\textbf{Gate}\\
\midrule
$f=$triple&$I_1(f)=1, I_2(f)=I_3(f)=I_4(f)=0,$\\
$f=$ground rule 1&$I_2(f)=1, I_1(f)=I_3(f)=I_4(f)=0,$\\
$f=$ground rule 2&$I_3(f)=1, I_1(f)=I_2(f)=I_4(f)=0,$\\
$f=$ground rule 3&$I_4(f)=1, I_1(f)=I_2(f)=I_3(f)=0,$\\
\bottomrule
\end{tabular}
\end{minipage}
\end{table}

If $f^+$ is a triple sample $(h, r, t)$, the $f^-$ is composed of the triple with either the head or tail replaced by a random entity (but not both at the same time), which is shown as the following equation:
\begin{equation}\label{eq:negative_triple}
\mathcal{N}_{f^+}=\{{(h',r,t|h'\in \mathcal{E})}\cup{(h,r,t'|t'\in \mathcal{E})}\}.
\end{equation}

If $f^+$ is a ground rule sample of rule 1 $(h, r_1, t)\Rightarrow(h, r_2, t)$,  the $f^-$ can be also constructed by replacing the head or tail with a random entity according to Equation (\ref{eq:negative_rule1}):
\begin{equation}\label{eq:negative_rule1}
\begin{split}
\mathcal{N}_{f^+}=\{{[(h', r_1, t)\Rightarrow(h', r_2, t)|h'\in \mathcal{E}]}\\
\cup{[(h, r_1, t')\Rightarrow(h, r_2, t')|t'\in \mathcal{E}]}\}.
\end{split}
\end{equation}

If $f^+$ is a ground rule sample of rule 2 $(e_1, r_1, e_2)+(e_2,r_2,e_3)$ $\Rightarrow(e_1, r_3, e_3)$,  the $f^-$ can be constructed by replacing the $e_1$ or $e_3$ with a random entity according to Equation (\ref{eq:negative_rule2}):
\begin{equation}\label{eq:negative_rule2}
\begin{split}
\mathcal{N}_{f^+}\!=\{{[(e_1', r_1, e_2)+(e_2,r_2,e_3)\!\Rightarrow\!(e_1', r_3, e_3)|e_1'\in \mathcal{E}]}\\
\!\cup{[(e_1, r_1, e_2)+(e_2,r_2,e_3')\!\Rightarrow\!(e_1, r_3, e_3')|e_3'\in \mathcal{E}]}\}.
\end{split}
\end{equation}

If $f^+$ is a ground rule sample of rule 3 $(h, r_1, t)\Leftrightarrow(t, r_2, h)$,  the $f^-$ can be also constructed by replacing the head or tail with a random entity according to Equation (\ref{eq:negative_rule3}):
\begin{equation}\label{eq:negative_rule3}
\begin{split}
\mathcal{N}_{f^+}=\{{[(h', r_1, t)\Leftrightarrow(t, r_2, h')|h'\in \mathcal{E}]}\\
\cup{[(h, r_1, t')\Leftrightarrow(t', r_2, h)|t'\in \mathcal{E}]}\}.
\end{split}
\end{equation}

All of the corrupted triples or ground rules ($f^-$) cannot exist in the original triples and ground rules. The loss function, shown in Equation (\ref{eq:loss}), favors lower values of the energy for training triples or rules than for corrupted triples or rules, and this is a natural implementation of the intended criterion. In this paper, the embedding representation of an entity is the same when the entity appears in the head position and tail position of a triple. The optimization is carried out by stochastic gradient descent, over the possible $\mathbf{h}$, $\mathbf{r}$ and $\mathbf{t}$.

\section{Discussions}
we discuss several practical issues of our proposed model.
\textbf{Uniqueness}. KALE \cite{Guo:emnlp2016} encode the rules in the form of true value. For example, KALE encode an inference ground rule $f\triangleq(h_m,r_1,t_n)\Rightarrow(h_m,r_2,t_n)$ using the following Equation:
\begin{equation}\label{eq:KALE_Rule}
I(f)=s(h_m,r_1,t_n)\cdot s(h_m,r_2,t_n)-s(h_m,r_1,t_n)+1.
\end{equation}
where $s()$ can be Equation \ref{eq:transe}, \ref{eq:transh}, or \ref{eq:transr}. However, this rule modeling method only consider the true value of a triple. Many different triples can generate a same score using $s()$. Thus, many incorrect rules can easily meet the Equation \ref{eq:KALE_Rule}. The knowledge graph embedding model would inference inaccurate results according this rule modeling method after learning. In contrast, to guarantee the encoding form of a rule and the rule have one-to-one mapping relation, we propose a general interaction operation for rules to make the components within rules directly interact in the vector space. As shown in Section 4.1, Section 4.2, and Section 4.3, each component interact with each other by the algebraic operation in the vector space, and then our rule modeling method return a true value.

\noindent\textbf{Consistency}. We propose to map all of the triples and rules into first-order logics to achieve the algebraic operations consistency. For a triple ($h$,$r$,$t$), the relation $r$ is regarded as a function operating on the head entity $h$ and the tail entity $t$ can be deduced after function operation in first-order logic. Thus, a triple ($h$,$r$,$t$) can be represented as $r(h)\Rightarrow t$. Similarly, we can also obtain the first-order logic forms of the ground rules as shown in Table \ref{tab:fol}. The common logic symbols between triples and rules are $\Rightarrow$ and $\wedge$, whose algebraic operations have been defined in Table 3. Thus, the triples and rules use the same set of algebraic operations. In contrast, KALE \cite{Guo:emnlp2016} use the Equation \ref{eq:KALE_Rule} to express the $\Rightarrow$ symbol in rule model and use subtraction operation in triple.

\section{Experiments}
In this section, we introduce the experimental design, datasets, and results of our proposed model. Through these experiments, we want to verify whether our rule-enhanced method can improve the performance of the translation based knowledge graph embedding models. We use a library for parallel knowledge graph embedding, ParaGraphE\footnote{https://github.com/LIBBLE/LIBBLE-MultiThread/tree/master/ParaGraphE} \cite{Niu:2017}, to run several baseline models and implement our model. The results of baselines may be slightly different from the published results due to the asynchronous update in	multi-thread setting.

\subsection{Datasets}

\textbf{Wordnet} \cite{Miller:1}: This knowledge base is designed to generate an intuitively usable combination of dictionary and thesaurus, and it is used in natural language processing tasks. Its entities (termed synsets) correspond to word senses, and relationships define lexical relations between them. We used the same data version used in \cite{Bordes:2011,Bordes:2013,Bordes:2014}: WN18, a subset of Wordnet. Examples of triples are ($\_score\_NN\_1$, $\_hypernym$, $\_evaluation\_NN\_1$) or ($\_score\_NN\_2$, $\_has\_part$, $\_musical\_notation\_NN\_1$).

\textbf{Freebase} \cite{Bollacker:1}: Freebase is a large collaborative and growing knowledge base of general facts. We use the same data version used in \cite{Bordes:2011,Bordes:2013}: FB15K, a relatively dense subgraph of Freebase where all entities are present in Wikilinks database. In addition, we follow KALE \cite{Guo:emnlp2016} to create a small data version from FB15K with a little difference from KALE. KALE keeps the triples whose relation types start from ``people'', ``location'', or ``sports'' in FB15K. After that, there are totally 192 such relations. \citet{Guo:emnlp2016} manually filtered some ambiguous relations, which is not introduced in detail. To approximate the dataset generated by \citet{Guo:emnlp2016}, we only keep relations whose object type of head entity and tail entity also start from ``people'', ``location'', and ``sports'' from above 192 relations. Finally, there are only 166 relations. This dataset is denoted as FB166 in the rest of this section.

The WN18 and FB15K are released in \cite{Bordes:2013}. For WN18 and FB15K we use the original data split, and for FB166 we extract triples associated with the 166 relations from the training, validation, and testing sets of FB15K. Table \ref{tab:data_statistic} gives more details of the three datasets, including the number of entities, relations, triples in train/valid/test set.

\begin{table}[t]
\begin{minipage}{1\linewidth}
\centering
\renewcommand{\multirowsetup}{\centering}
\renewcommand\arraystretch{1}
\caption{Statistic of datasets.} \label{tab:data_statistic}
\begin{tabular}{cccccc}
\toprule
\textbf{Dataset}&\textbf{\#E}&\textbf{\#R}&\multicolumn{3}{c}{\textbf{\#Trip.}(Train / Valid / Test)}\\
\midrule
FB15K&14,951&1,345&483,142&50,000&59,071\\
FB166&9,658&166&100,289&10,457&12,327\\
WN18&40,943&18&141,442&5,000&5,000\\
\bottomrule
\end{tabular}
\end{minipage}
\end{table}

We further construct the three types of logic rules for each dataset using the method described in Section 3. Table \ref{tab:rule_statistic} shows the details of the mined rules: rule 1 (inference rule), rule 2 (transitivity rule), and rule 3 (antisymmetry rule). For the FB166 and FB15K dataset, the thresholds $\tau$ of the rule 1, rule 2 and rule 3 are set as 0.5, 0.6, and 0.5 respectively. For the WN18 dataset, the thresholds $\tau$ of the rule 1, rule 2 and rule 3 are set as 0.5, 0.5, 0.5 respectively. All of the thresholds are selected according to the validation set. After mining the logical rules, the mined rules are then instantiated with concrete entities (grounding) using the method described in Section 3. The statistics of ground rules are shown in Table \ref{tab:gound_rule_statistic}. Table \ref{tab:Exa_rules} shows some examples of the rules in three data sets.

\begin{table}[t]
\begin{minipage}{1\linewidth}
\centering
\renewcommand{\multirowsetup}{\centering}
\renewcommand\arraystretch{1}
\caption{The rule statistic.} \label{tab:rule_statistic}
\begin{tabular}{ccccc}
\toprule
\textbf{Dataset}&\textbf{\#Rule 1}&\textbf{\#Rule 2}&\textbf{\#Rule 3}&\textbf{Total}\\
\midrule
FB15K&160&2537&637&3334\\
FB166&38&157&73&268\\
WN18&0&0&11&11\\
\bottomrule
\end{tabular}
\end{minipage}
\end{table}

\begin{table}[t]
\begin{minipage}{1\linewidth}
\centering
\renewcommand{\multirowsetup}{\centering}
\renewcommand\arraystretch{1}
\caption{The ground rule statistic} \label{tab:gound_rule_statistic}
\begin{tabular}{ccccc}
\toprule
\textbf{Dataset}&\textbf{\#Rule 1}&\textbf{\#Rule 2}&\textbf{\#Rule 3}&\textbf{Total}\\
\midrule
FB15K&8,508&152,987&113,476&274,971\\
FB166&2,254&20,444&20,946&43,644\\
WN18&0&0&9,352&9,352\\
\bottomrule
\end{tabular}
\end{minipage}
\end{table}

\begin{table*}[ht]
\begin{minipage}{1\linewidth}
\centering
\renewcommand{\multirowsetup}{\centering}
\renewcommand\arraystretch{1.1}
\caption{Examples of rules created.} \label{tab:Exa_rules}
\begin{tabular}{|c|m{15.5cm}|}
\hline
\textbf{Dataset}&\textbf{Rule exmaples}\\
\hline
\multirow{5}{*}{FB}
&$\forall x,y:$ /location/country/capital$(x,y)$ $\Rightarrow$ /location/location/contains$(x,y)$\\
\cline{2-2}
&$\forall x,y:$ /location/country/first$\_$level$\_$divisions$(x,y)$ $\Rightarrow$ /location/location/contains$(x,y)$\\
\cline{2-2}
&$\forall x,y,z:$ /people/person/place$\_$of$\_$birth$(x,y)$ + /location/administrative$\_$division/second$\_$level$\_$division$\_$of$(y,z)$ $\Rightarrow$ /people/person/nationality$(x,z)$\\
\cline{2-2}
&$\forall x,y:$ /people/family/members$(x,y)$ $\Leftrightarrow$ /people/familymember/family$(y,x)$\\
\cline{2-2}
&$\forall x,y:$ /location/location/people$\_$born$\_$here$(x,y)$ $\Leftrightarrow$ /people/person/place$\_$of$\_$birth$(y,x)$\\
\hline
\multirow{4}{*}{WN18}
&$\forall x,y:$ $\_$member$\_$of$\_$domain$\_$usage$(x,y)$ $\Leftrightarrow$ $\_$synset$\_$domain$\_$usage$\_$of$(y,x)$\\
\cline{2-2}
&$\forall x,y:$ $\_$hyponym$(x,y)$ $\Leftrightarrow$ $\_$hypernym$(y,x)$\\
\cline{2-2}
&$\forall x,y:$ $\_$part$\_$of$(x,y)$ $\Leftrightarrow$ $\_$has$\_$part$(y,x)$\\
\cline{2-2}
&$\forall x,y:$ $\_$instance$\_$hypernym$(x,y)$ $\Leftrightarrow$ $\_$instance$\_$hyponym$(y,x)$\\
\hline
\end{tabular}
\end{minipage}
\end{table*}

\subsection{Link Prediction}
This task is designed to complete a triple ($h,r,t$) with $h$ or $t$ missing. Instead of requiring the best answer, this task focuses on the ranking of candidate entities from the knowledge.

\textbf{Evaluation protocol}: We use the same metric as TransE \cite{Bordes:2013}: for each testing triple ($h, r, t$), we replace the head entity $h$ by every entity $e$ in the knowledge graph and calculate a dissimilarity score (by the Equation (\ref{eq:transe}), (\ref{eq:transh}), or (\ref{eq:transr})) on the corrupted triple ($e, r, t$). The rank of the original correct triple is obtained after ranking the scores in ascending order. Similarly, we can get another rank for ($h, r, t$) by corrupting the tail $t$.Aggregating over all the testing triples $\mathcal{K}_t$, we use three metrics to do the evaluation:

\begin{enumerate}
  \item \textbf{MR}: the value of averaged rank or Mean Rank. The smaller, the better. MR is calculated by:
        \begin{equation}\label{eq:MR}
        MR=\frac{1}{2\#\mathcal{K}_t}\sum_{i=1}^{\#\mathcal{K}_t}{(rank_{ih}+rank_{it})},
        \end{equation}
        where $rank_{ih}$ and $rank_{it}$ refers to the rank position of the head correct triple for the $i^{th}$ triple by corrupting the head and tail entity.
  \item \textbf{MRR}: the value of mean reciprocal rank. The higher, the better. The reciprocal rank is the multiplicative inverse of the rank of the correct triple. The mean reciprocal rank is the average of the reciprocal ranks of results for all triples. MRR is calculated by:
        \begin{equation}\label{eq:MRR}
        MRR=\frac{1}{2\#\mathcal{K}_t}\sum_{i=1}^{\#\mathcal{K}_t}(\frac{1}{rank_{ih}}+\frac{1}{rank_{it}}),
        \end{equation}
        where $rank_{ih}$ and $rank_{it}$ refers to the same meaning as introducing in MR metric.
  \item \textbf{Hits@n}: the proportion of rank not larger than n. The higher, the better. Hits@10, Hits@5, Hits@3, and Hits@1 are used in our experiments. Hits@n is calculated by:
      \begin{equation}\label{eq:hits}
        Hits@n=\frac{1}{2\#\mathcal{K}_t}\sum_{i=1}^{\#\mathcal{K}_t}(I_n(rank_{ih})+I_n(rank_{it})),
      \end{equation}
      \begin{equation}
        I_n(rank_i)=\left\{\begin{array}{ll}
        1 & \textrm{if $rank_i\leq n$},\\
        0 & \textrm{otherwise}.
        \end{array} \right.
      \end{equation}
\end{enumerate}

We name original setting as ``\emph{Raw}'' setting. If the newly created triple really exists in the knowledge graph, ranking it before the original triple is not wrong. For example, both of a testing triple {\fontfamily{qcr}\selectfont (Pairs}, {\fontfamily{qcr}\selectfont Located-In}, {\fontfamily{qcr}\selectfont France)} and a possible corruption {\fontfamily{qcr}\selectfont (Lyon}, {\fontfamily{qcr}\selectfont Located-In}, {\fontfamily{qcr}\selectfont France)} are valid. In this case, ranking {\fontfamily{qcr}\selectfont Lyon} before the correct answer {\fontfamily{qcr}\selectfont Pairs} should not be counted as an error. Thus, we take the same filtering operation used in \cite{Bordes:2013} to filter out these triples in training, validation and testing data, which is named as ``\emph{Filt.}'' setting. Finally, we remove {\fontfamily{qcr}\selectfont Lyon} from the candidate entity list before obtaining the rank of {\fontfamily{qcr}\selectfont Pairs} in the above exmaple.

\textbf{Implementation}: As FB15K and WN18 are also used in previous work, the parameters of TransE \cite{Bordes:2013}, TransH \cite{Wang:2014a}, and TransR \cite{Lin:2015a} are set as the same with the parameters in the published works except that the margin $\gamma$ of TransH is set to 2 and 3 on FB15K and WN18 respectively. Because this configuration achieves the best results in validation set. For the FB166 dataset, we set the same parameters for TransE, TransH, and TransR with the parameters used in FB15K except that the margin $\gamma$ of TransH is set to 3. The parameters of our proposed rule-enhanced TransE, TransH, and TransR on three datasets are set as the same with the parameters used in original TransE, TransH and TransR except for the learning rate $\alpha$ on FB15K. The learning rate $\alpha$ of the rule-enhanced TransE, TransH, and TransR on FB15K are set as 0.001, 0.004, and 0.0001 respectively. To avoid overfitting, we initialize the entity and relation embeddings with the results of each enhanced model trained on triples, inference rules and transitivity rules. Based on the initialized entity and relation embeddings, we train the enhanced model on antisymmetry rules by using the new learning rate. Here, the ``unif'' sampling is used in corrupting negative sample. The optimal parameters are determined by the validation set.


\begin{table*}[t]
\begin{minipage}{1\linewidth}
\centering
\renewcommand{\multirowsetup}{\centering}
\renewcommand\arraystretch{1}
\caption{Link prediction results on dataset FB166.} \label{tab:lr_FB166}
\begin{tabular}{@{ }c@{ }@{ }c@{ }@{ }c@{ }@{ }c@{ }@{ }c@{ }@{ }c@{ }@{ }c@{ }@{ }c@{ }@{ }c@{ }@{ }c@{ }@{ }c@{ }@{ }c@{ }@{ }c@{ }}
\toprule
\multirow{2}{*}{\bf Method}
&\multicolumn{2}{c}{\textbf{MR}}&\multicolumn{2}{c}{\textbf{MRR}}&\multicolumn{2}{c}{\textbf{Hits@10}}&\multicolumn{2}{c}{\textbf{Hits@5}}&\multicolumn{2}{c}{\textbf{Hits@3}}&\multicolumn{2}{c}{\textbf{Hits@1}}\\
\cmidrule(r){2-3}\cmidrule(r){4-5}\cmidrule(r){6-7}\cmidrule(r){8-9}\cmidrule(r){10-11}\cmidrule(r){12-13}
&Raw&Filt.&Raw&Filt.&Raw&Filt.&Raw&Filt.&Raw&Filt.&Raw&Filt.\\
\midrule
TransE&523&229&0.1679&0.3226&0.3871&0.5707&0.2942&0.4850&0.2083&0.4108&0.0633&0.1825\\
TransE(Pre)&428&232&0.2426&0.4369&0.4266&0.6586&0.3522&0.5696&0.2769&0.5753&0.1100&0.2938\\
TransE(Rule)&\textbf{408}&\textbf{210}&\textbf{0.3413}&\textbf{0.6584}&\textbf{0.5098}&\textbf{0.7392}&\textbf{0.4533}&\textbf{0.7111}&\textbf{0.3996}&\textbf{0.6897}&\textbf{0.2455}&\textbf{0.6100}\\
\midrule
TransH&431&\textbf{191}&0.2127&0.3332&0.4168&0.5429&0.3358&0.4629&0.2596&0.3970&0.1097&0.2183\\
TransH(Pre)&415&194&0.1539&0.2033&0.3812&0.4951&0.2845&0.3731&0.2073&0.2879&0.0362&0.0449\\
TransH(Rule)&\textbf{407}&280&\textbf{0.3251}&\textbf{0.5818}&\textbf{0.5126}&\textbf{0.7325}&\textbf{0.4523}&\textbf{0.6969}&\textbf{0.3983}&\textbf{0.6640}&\textbf{0.2145}&\textbf{0.4802}\\
\midrule
TransR&499&185&0.1928&0.4194&0.4344&0.6484&0.3319&0.5711&0.2306&0.5055&0.0839&0.2869\\
TransR(Pre)&395&218&0.2768&0.4424&0.4457&0.6066&0.3717&0.5299&0.3093&0.4715&\textbf{0.2033}&0.3627\\
TransR(Rule)&\textbf{392}&\textbf{152}&\textbf{0.2835}&\textbf{0.6464}&\textbf{0.5011}&\textbf{0.7849}&\textbf{0.4204}&\textbf{0.7486}&\textbf{0.3343}&\textbf{0.7104}&0.1642&\textbf{0.5615}\\
\midrule
KALE&510&217&0.1897&0.3819&0.4115&0.6061&0.3103&0.5057&0.2247&0.4198&0.0854&0.1947\\
\bottomrule
\end{tabular}
\end{minipage}
\end{table*}

\begin{table*}[t]
\begin{minipage}{1\linewidth}
\centering
\renewcommand{\multirowsetup}{\centering}
\renewcommand\arraystretch{1}
\caption{Link prediction results on dataset FB15K.} \label{tab:lr_FB15K}
\begin{tabular}{@{ }c@{ }@{ }c@{ }@{ }c@{ }@{ }c@{ }@{ }c@{ }@{ }c@{ }@{ }c@{ }@{ }c@{ }@{ }c@{ }@{ }c@{ }@{ }c@{ }@{ }c@{ }@{ }c@{ }}
\toprule
\multirow{2}{*}{\bf Method}
&\multicolumn{2}{c}{\textbf{MR}}&\multicolumn{2}{c}{\textbf{MRR}}&\multicolumn{2}{c}{\textbf{Hits@10}}&\multicolumn{2}{c}{\textbf{Hits@5}}&\multicolumn{2}{c}{\textbf{Hits@3}}&\multicolumn{2}{c}{\textbf{Hits@1}}\\
\cmidrule(r){2-3}\cmidrule(r){4-5}\cmidrule(r){6-7}\cmidrule(r){8-9}\cmidrule(r){10-11}\cmidrule(r){12-13}
&Raw&Filt.&Raw&Filt.&Raw&Filt.&Raw&Filt.&Raw&Filt.&Raw&Filt.\\
\midrule
TransE&237&121&0.1575&0.2466&0.3404&0.4646&0.2376&0.3577&0.1711&0.2829&0.0698&0.1376\\
TransE(Pre)&195&96&0.1872&0.3067&0.3844&0.5332&0.2774&0.4261&0.2073&0.3506&0.0922&0.1931\\
TransE(Rule)&\textbf{163}&\textbf{72}&\textbf{0.2923}&\textbf{0.5688}&\textbf{0.5403}&\textbf{0.7779}&\textbf{0.4273}&\textbf{0.7075}&\textbf{0.3404}&\textbf{0.6450}&\textbf{0.1709}&\textbf{0.4517}\\
\midrule
TransH&227&116&0.1834&0.2827&0.3759&0.5125&0.2706&0.4064&0.2023&0.3322&0.0907&0.1729\\
TransH(Pre)&235&136&0.1131&0.1572&0.2945&0.3884&0.1916&0.2784&0.1279&0.1985&0.0232&0.0365\\
TransH(Rule)&\textbf{175}&\textbf{106}&\textbf{0.2870}&\textbf{0.5170}&\textbf{0.5475}&\textbf{0.7696}&\textbf{0.4328}&\textbf{0.6974}&\textbf{0.3423}&\textbf{0.6207}&\textbf{0.1582}&\textbf{0.3708}\\
\midrule
TransR&218&72&0.2039&0.3774&0.4385&0.6440&0.3131&0.5332&0.2277&0.4455&0.0967&0.2400\\
TransR(Pre)&236&153&0.1774&0.2764&0.3165&0.4432&0.2323&0.3521&0.1820&0.2958&0.1084&0.1931\\
TransR(Rule)&\textbf{175}&\textbf{55}&\textbf{0.2380}&\textbf{0.4646}&\textbf{0.4724}&\textbf{0.7088}&\textbf{0.3529}&\textbf{0.6119}&\textbf{0.2680}&\textbf{0.5335}&\textbf{0.1284}&\textbf{0.3366}\\
\midrule
KALE&221&124&0.1864&0.2957&0.3725&0.4956&0.2681&0.3614&0.1841&0.3091&0.0702&0.1491\\
\bottomrule
\end{tabular}
\end{minipage}
\end{table*}

\begin{table*}[t]
\begin{minipage}{1\linewidth}
\centering
\renewcommand{\multirowsetup}{\centering}
\renewcommand\arraystretch{1}
\caption{Link prediction results on dataset WN18.} \label{tab:lr_WN18}
\begin{tabular}{@{ }c@{ }@{ }c@{ }@{ }c@{ }@{ }c@{ }@{ }c@{ }@{ }c@{ }@{ }c@{ }@{ }c@{ }@{ }c@{ }@{ }c@{ }@{ }c@{ }@{ }c@{ }@{ }c@{ }}
\toprule
\multirow{2}{*}{\bf Method}
&\multicolumn{2}{c}{\textbf{MR}}&\multicolumn{2}{c}{\textbf{MRR}}&\multicolumn{2}{c}{\textbf{Hits@10}}&\multicolumn{2}{c}{\textbf{Hits@5}}&\multicolumn{2}{c}{\textbf{Hits@3}}&\multicolumn{2}{c}{\textbf{Hits@1}}\\
\cmidrule(r){2-3}\cmidrule(r){4-5}\cmidrule(r){6-7}\cmidrule(r){8-9}\cmidrule(r){10-11}\cmidrule(r){12-13}
&Raw&Filt.&Raw&Filt.&Raw&Filt.&Raw&Filt.&Raw&Filt.&Raw&Filt.\\
\midrule
TransE&263&251&0.2803&0.3627&0.7256&0.8447&0.5629&0.7148&0.4131&0.5771&0.0564&0.086\\
TransE(Pre)&236&228&0.3331&0.4234&0.7914&0.8977&0.6610&0.8125&0.5137&0.6882&0.0786&0.1177\\
TransE(Rule)&\textbf{231}&\textbf{227}&\textbf{0.7104}&\textbf{0.8189}&\textbf{0.9164}&\textbf{0.9484}&\textbf{0.8893}&\textbf{0.9423}&\textbf{0.8562}&\textbf{0.9351}&\textbf{0.5550}&\textbf{0.7013}\\
\midrule
TransH&314&\textbf{298}&0.2680&0.3611&0.7546&0.8878&0.5930&0.7764&0.4223&0.6243&0.0250&0.0475\\
TransH(Pre)&326&316&0.2775&0.3475&0.7913&0.9031&0.6446&0.8023&0.4740&0.6508&0.0014&0.0018\\
TransH(Rule)&\textbf{310}&310&\textbf{0.4370}&\textbf{0.5041}&\textbf{0.9025}&\textbf{0.9439}&\textbf{0.8529}&\textbf{0.9290}&\textbf{0.7775}&\textbf{0.8966}&\textbf{0.0861}&\textbf{0.1201}\\
\midrule
TransR&226&213&0.3307&0.4629&0.7876&0.9295&0.6551&0.8720&0.5034&0.7605&0.0853&0.1428\\
TransR(Pre)&\textbf{212}&\textbf{210}&0.6512&0.7561&0.8928&0.9415&0.8402&0.9203&0.7599&0.8775&\textbf{0.5074}&0.6244\\
TransR(Rule)&233&228&\textbf{0.654}&\textbf{0.7863}&\textbf{0.9068}&\textbf{0.9455}&\textbf{0.8497}&\textbf{0.9286}&\textbf{0.7893}&\textbf{0.9108}&0.4984&\textbf{0.6594}\\
\midrule
KALE&245&216&0.4203&0.5727&0.7214&0.8659&0.6029&0.7248&0.4731&0.6371&0.0847&0.1463\\
\bottomrule
\end{tabular}
\end{minipage}
\end{table*}

\textbf{Results}: The results on FB166, FB15K, and WN18 datasets are shown in Table \ref{tab:lr_FB166}, Table \ref{tab:lr_FB15K}, and Table \ref{tab:lr_WN18} respectively. KALE was proposed by \citet{Guo:emnlp2016}. The TransE(Pre), TransH(Pre), and TransR(Pre) denote that the triples inferred from rules are used directly as training instances in TransE, TransH, and TransR. The TransE(Rule), TransH(Rule), and TransR(Rule) denote that the rules are jointly used with the original triples. From the results of three datasets, we can see that: (1) The Pre version of TransE, TransH, and TransR models outperform the original TransE, TransH, TransR models which use triples alone on three datasets in most cases. This implies that our rule extraction method can generate correct rules, which can infer new triples. (2) The logic rule-enhanced TransE, TransH, and TransR models greatly improve the performance of the original TransE, TransH, TransR methods which use triples alone. This implies that our rule-enhanced method can efficiently help the translation based knowledge graph embedding models to learn better knowledge representation, e.g., entity embedding and relation embedding. In addition, our rule-enhanced method can encode more information of a rule compared with adding an inferred triple into training data. (3) Our rule-enhanced TransE model outperforms the KALE model \cite{Guo:emnlp2016}. The KALE just encodes rules into TransE model. We implement it using the ParaGraphE. (4) Our rule-enhanced method significantly improves results on the Hits@1 metric, e.g., the Hits@1 increases from 0.086 to 0.7013 on WN18 dataset using rule-enhanced TransE model. The first rank metric is the most important in real knowledge inference task. The Hits@1 score 0.086 indicates that the original TransE model almost cannot rank the the correct triple in the first. Thus, our method can make knowledge graph embedding method capable of being used in the real knowledge inference task. On the whole, our rule-enhanced method can efficiently leverage the rules to embed the entities and relations into a low-dimensional space.

\subsection{Triple Classification}
The task of triple classification is to confirm whether a given triple ($h, r, t$) is correct or not, i.e., binary classification on a triple. It is used in \cite{Wang:2014a} to evaluate TransH model.

\textbf{Evaluation protocol}: We follow the same protocol in NTN \cite{Socher:2013} and TransH \cite{Wang:2014a}. Evaluation of classification needs negative labels. Thus, we first create labeled data for evaluation. For each triple in the testing or validation set (i.e., a positive triple), we construct 10 negative triples for it by randomly corrupting the entities, 5 at the head position and the other 5 at the tail position\footnote{Previous works\cite{Socher:2013,Wang:2014a} always construct only one negative triple for each testing triple or validation triple. To increase the classification difficulty, we try to use a highly unbalanced setting, with a positive-to-negative ratio of 1:10.}. To further make the negative triples as difficult as possible, we corrupt a position using only entities that have appeared in the position. For example, to construct a corrupted triple given a triple ($h_o, r, t$) by corrupting head entity, the randomly selected head entity $h_c$ must be linked to another tail entity by the relation $r$. In addition, we need to ensure that each corrupted triple does not exist in either the training, validation, or testing set.

The decision rule for classification is simple: for a triple ($h, r, t$), we simply use the dissimilarity score (by the Equation (\ref{eq:transe}), (\ref{eq:transh}), or (\ref{eq:transr})) to classify triples. If the dissimilarity score is below a relation-specific threshold $\sigma_r$, the triple is classified into positive. Otherwise negative. The relation-specific threshold $\sigma_r$ is set according to the validation set.

\textbf{Implementation}: In the triple classification, the datasets are the same as the datasets used in link prediction task. The search space of hyperparameters is identical to link prediction. Thus, the optimal hyperparameters are set the same as the hyperparameters in the link prediction task.

\textbf{Results}:
The accuracies of the triple classification are shown in Table \ref{tab:tc}. From the results, we can see that: (1) The Per-version based TransE and TransH slightly outperform the original models on the three datasets in most cases. (2) Our rule-enhanced method outperforms other models including the original models, the Per-version based model and the KALE model. This once again validates that our method can generate high-quality rules and better leverage the rules in the knowledge graph embedding than baselines.

\begin{table}[t]
\begin{minipage}{1\linewidth}
\centering
\renewcommand{\multirowsetup}{\centering}
\renewcommand\arraystretch{1}
\caption{Accuracies of triple classification on the testing dataset of FB166, FB15K, and WN18.} \label{tab:tc}
\begin{tabular}{cccc}
\toprule
Dataset&FB166&FB15K&WN18\\
\midrule
TransE&0.9106&0.9094&0.9580\\
TransE(Per)&0.9097&0.9096&0.9601\\
TransE(Rule)&\textbf{0.9490}&\textbf{0.9256}&\textbf{0.9873}\\
\midrule
TransH&0.9091&0.9086&0.9470\\
TransH(Per)&0.9093&0.9090&0.9513\\
TransH(Rule)&\textbf{0.9505}&\textbf{0.9192}&\textbf{0.9936}\\
\midrule
TransR&0.9146&0.9091&0.9523\\
TransR(Per)&0.8727&0.8258&0.9686\\
TransR(Rule)&\textbf{0.9204}&\textbf{0.9208}&\textbf{0.9926}\\
\midrule
KALE&0.9257&0.9011&0.9779\\
\bottomrule
\end{tabular}
\end{minipage}
\end{table}

\section{Related Work}
Recently, many works have made great efforts on modeling knowledge graphs. Knowledge graph embedding encodes entities of a knowledge graph into a continuous low-dimensional space as vectors, and encodes relations as vectors or matrices. Several works use neural network architectures \cite{Socher:2013,Bordes:2014}, tensor factorization \cite{Nickel:2011,Riedel:2013,Chang:2014}, and Bayesian clustering strategies \cite{Xu:2006,Sutskever:2009} to explain triples via latent representation of entities and relations. To capture both the first-order and the second-order interactions between two entities, \citet{Jenatton:2012} proposed Latent Factor Model (LFM) to adopt a bilinear function as its score function. Recently, TransE \cite{Bordes:2013}, which models the relation as a translation from head entity to tail entity, achieves a good trade-off between prediction accuracy and computational efficiency. TransE is efficient when modeling simple relations, such as one-to-one relations. To improve the performance of TransE on complicated relations, various extensions like TransH \cite{Wang:2014a}, TransR \cite{Lin:2015a}, and TransD \cite{Ji:1} are proposed. TransH allows entities to have multiple representations. To obtain multiple representations of an entity, TransH projects an entity vector into relation-specific hyperplanes. TransR also handles the problem of TransE by introducing relation spaces. It allows an entity to have various vector representations by mapping an entity vector into relation-specific spaces. To handle multiple types of relations, TransD constructs relation mapping matrices dynamically by considering entities and a relation simultaneously. However, these models lose the simplicity and efficiency of TransE. To combine the power of tensor product with the efficiency and simplicity of TransE, HOLE \cite{Nickel:2015} uses the circular correlation of vectors to represent pairs of entities.

Most existing methods embed the knowledge graph based solely on triples contained in the knowledge graph. Several recent works try to incorporate other available knowledge, e.g., relation paths \cite{Lin:2015b,Luo:2015}, entity types \cite{Guo:2015}, Wikipedia anchors \cite{Wang:2014b}, and entity descriptions \cite{Zhong:2015}. Logic rules have been widely used in knowledge inference and acquisition \cite{Brocheler:2010,Pujara:2013}. Recent works put growing interest in logic rules. \citet{Wang:2015} and \citet{Wei:2015} tried to utilize rules via integer linear programming or Markov logic networks. However, rules are modeled separately from embedding models and cannot help in obtaining better embeddings in these approaches. \citet{Rocktaschel:2015} proposed a joint model which jointly encodes the rules into the embedding. However, this work focuses on relation extraction task and learns the entity pair embedding instead of entity embeddings. In \cite{Guo:emnlp2016} a new method named KALE is proposed to jointly embed the knowledge graph and logic rules. The rules are represented as complex formulae and are modeled by the t-norm fuzzy logics.

\section{Conclusion and Future Work}
In conclusion, we present a novel logic rule-enhanced method which can be easily integrated with any translation based knowledge graph embedding model. Besides that, an automatic logic rule extraction method is introduced in this work. Our method places the logic rules and triples in the same space. Our experiments show that our proposed method can efficiently improve the performance of knowledge graph embedding in various tasks. Particularly, it brings a great improvement in the filtered Hits@1 evaluation which is a pivotal evaluation in the knowledge inference task. This makes the knowledge graph embedding method capable being applied in the real and large scale knowledge inference tasks.

This is our first step in the scope of external knowledge powered knowledge graph embedding. In the future, we will explore more types of effective rules that exist in data. Besides, we plan to handle the embedding of multiple knowledge graphs, such as jointly embedding Freebase and YAGO.

%% file: encode_rule_into_knowledge_graph_embedding.bbl

\begin{thebibliography}{00}


\ifx \showCODEN    \undefined \def \showCODEN     #1{\unskip}     \fi
\ifx \showDOI      \undefined \def \showDOI       #1{#1}\fi
\ifx \showISBNx    \undefined \def \showISBNx     #1{\unskip}     \fi
\ifx \showISBNxiii \undefined \def \showISBNxiii  #1{\unskip}     \fi
\ifx \showISSN     \undefined \def \showISSN      #1{\unskip}     \fi
\ifx \showLCCN     \undefined \def \showLCCN      #1{\unskip}     \fi
\ifx \shownote     \undefined \def \shownote      #1{#1}          \fi
\ifx \showarticletitle \undefined \def \showarticletitle #1{#1}   \fi
\ifx \showURL      \undefined \def \showURL       {\relax}        \fi
\providecommand\bibfield[2]{#2}
\providecommand\bibinfo[2]{#2}
\providecommand\natexlab[1]{#1}
\providecommand\showeprint[2][]{arXiv:#2}

\bibitem[\protect\citeauthoryear{Bollacker, Evans, Paritosh, Sturge, and
  Taylor}{Bollacker et~al\mbox{.}}{2008}]%
        {Bollacker:1}
\bibfield{author}{\bibinfo{person}{Kurt Bollacker}, \bibinfo{person}{Colin
  Evans}, \bibinfo{person}{Praveen Paritosh}, \bibinfo{person}{Tim Sturge},
  {and} \bibinfo{person}{Jamie Taylor}.} \bibinfo{year}{2008}\natexlab{}.
\newblock \showarticletitle{Freebase: a collaboratively created graph database
  for structuring human knowledge}.
\newblock \bibinfo{journal}{{\em In Proceedings of the 2008 ACM SIGMOD
  international conference on Management of data\/}} (\bibinfo{year}{2008}),
  \bibinfo{pages}{1247--1250}.
\newblock


\bibitem[\protect\citeauthoryear{Bordes, Glorot, Weston, and Bengio}{Bordes
  et~al\mbox{.}}{2014}]%
        {Bordes:2014}
\bibfield{author}{\bibinfo{person}{Antoine Bordes}, \bibinfo{person}{Xavier
  Glorot}, \bibinfo{person}{Jason Weston}, {and} \bibinfo{person}{Yoshua
  Bengio}.} \bibinfo{year}{2014}\natexlab{}.
\newblock \showarticletitle{A Semantic Matching Energy Function for Learning
  with Multi-relational Data}.
\newblock \bibinfo{journal}{{\em Machine Learning\/}} (\bibinfo{year}{2014}),
  \bibinfo{pages}{233--259}.
\newblock


\bibitem[\protect\citeauthoryear{Bordes, Usunier, Garcia-Dur\'an, Weston, and
  Yakhnenko}{Bordes et~al\mbox{.}}{2013}]%
        {Bordes:2013}
\bibfield{author}{\bibinfo{person}{Antoine Bordes}, \bibinfo{person}{Nicolas
  Usunier}, \bibinfo{person}{Alberto Garcia-Dur\'an}, \bibinfo{person}{Jason
  Weston}, {and} \bibinfo{person}{Oksana Yakhnenko}.}
  \bibinfo{year}{2013}\natexlab{}.
\newblock \showarticletitle{Translating embeddings for modeling multirelational
  data}.
\newblock \bibinfo{journal}{{\em In Advances in Neural Information Processing
  Systems\/}} (\bibinfo{year}{2013}), \bibinfo{pages}{2787--2795}.
\newblock


\bibitem[\protect\citeauthoryear{Bordes, Weston, Collobert, and Bengio}{Bordes
  et~al\mbox{.}}{2011}]%
        {Bordes:2011}
\bibfield{author}{\bibinfo{person}{Antoine Bordes}, \bibinfo{person}{Jason
  Weston}, \bibinfo{person}{Ronan Collobert}, {and} \bibinfo{person}{Yoshua
  Bengio}.} \bibinfo{year}{2011}\natexlab{}.
\newblock \showarticletitle{Learning structured embeddings of knowledge bases}.
\newblock \bibinfo{journal}{{\em In Proceedings of the Twenty-Fifth AAAI
  Conference on Artificial Intelligence\/}} (\bibinfo{year}{2011}),
  \bibinfo{pages}{301--306}.
\newblock


\bibitem[\protect\citeauthoryear{Br\"ocheler, Mihalkova, and
  Getoor}{Br\"ocheler et~al\mbox{.}}{2014}]%
        {Brocheler:2010}
\bibfield{author}{\bibinfo{person}{Matthias Br\"ocheler},
  \bibinfo{person}{Lilyana Mihalkova}, {and} \bibinfo{person}{Lise Getoor}.}
  \bibinfo{year}{2014}\natexlab{}.
\newblock \showarticletitle{Probabilistic similarity logic}.
\newblock \bibinfo{journal}{{\em In Proceedings of the 26th Conference on
  Uncertainty in Artificial Intelligence\/}} (\bibinfo{year}{2014}),
  \bibinfo{pages}{73--82}.
\newblock


\bibitem[\protect\citeauthoryear{Chang, Yih, Yang, and Meek}{Chang
  et~al\mbox{.}}{2014}]%
        {Chang:2014}
\bibfield{author}{\bibinfo{person}{Kai-Wei Chang}, \bibinfo{person}{Wen-Tau
  Yih}, \bibinfo{person}{Bishan Yang}, {and} \bibinfo{person}{Chris Meek}.}
  \bibinfo{year}{2014}\natexlab{}.
\newblock \showarticletitle{Typed tensor decomposition of knowledge bases for
  relation extraction}.
\newblock \bibinfo{journal}{{\em In Proceedings of the 2014 Conference on
  Empirical Methods in Natural Language Processing\/}} (\bibinfo{year}{2014}).
\newblock


\bibitem[\protect\citeauthoryear{Guo, Wang, Wang, Wang, and Guo}{Guo
  et~al\mbox{.}}{2015}]%
        {Guo:2015}
\bibfield{author}{\bibinfo{person}{Shu Guo}, \bibinfo{person}{Quan Wang},
  \bibinfo{person}{Bin Wang}, \bibinfo{person}{Lihong Wang}, {and}
  \bibinfo{person}{Li Guo}.} \bibinfo{year}{2015}\natexlab{}.
\newblock \showarticletitle{Semantically smooth knowledge graph embedding}.
\newblock \bibinfo{journal}{{\em In Proceedings of the 53rd Annual Meeting of
  the Association for Computational Linguistics and the 7th International Joint
  Conference on Natural Language Processing\/}} (\bibinfo{year}{2015}),
  \bibinfo{pages}{84--94}.
\newblock


\bibitem[\protect\citeauthoryear{Guo, Wang, Wang, Wang, and Guo}{Guo
  et~al\mbox{.}}{2016}]%
        {Guo:emnlp2016}
\bibfield{author}{\bibinfo{person}{Shu Guo}, \bibinfo{person}{Quan Wang},
  \bibinfo{person}{Lihong Wang}, \bibinfo{person}{Bin Wang}, {and}
  \bibinfo{person}{Li Guo}.} \bibinfo{year}{2016}\natexlab{}.
\newblock \showarticletitle{Jointly Embedding Knowledge Graphs and Logical
  Rules.}
\newblock \bibinfo{journal}{{\em In Proceedings of the 2016 Conference on
  Empirical Methods in Natural Language Processing\/}} (\bibinfo{year}{2016}),
  \bibinfo{pages}{192--202}.
\newblock


\bibitem[\protect\citeauthoryear{Hu, Ma, Liu, Hovy, and Xing}{Hu
  et~al\mbox{.}}{2016}]%
        {Hu:1}
\bibfield{author}{\bibinfo{person}{Zhiting Hu}, \bibinfo{person}{Xuezhe Ma},
  \bibinfo{person}{Zhengzhong Liu}, \bibinfo{person}{Eduard Hovy}, {and}
  \bibinfo{person}{Eric Xing}.} \bibinfo{year}{2016}\natexlab{}.
\newblock \showarticletitle{Harnessing deep neural networks with logic rules}.
\newblock \bibinfo{journal}{{\em arXiv preprint arXiv:1603.06318.\/}}
  (\bibinfo{year}{2016}).
\newblock


\bibitem[\protect\citeauthoryear{Jenatton, Roux, Bordes, and
  Obozinski}{Jenatton et~al\mbox{.}}{2012}]%
        {Jenatton:2012}
\bibfield{author}{\bibinfo{person}{Rodolphe Jenatton},
  \bibinfo{person}{Nicolas~Le Roux}, \bibinfo{person}{Antoine Bordes}, {and}
  \bibinfo{person}{Guillaume Obozinski}.} \bibinfo{year}{2012}\natexlab{}.
\newblock \showarticletitle{A latent factor model for highly multi-relational
  data.}
\newblock \bibinfo{journal}{{\em In Proceedings of the 25th International
  Conference on Neural Information Processing Systems\/}}
  (\bibinfo{year}{2012}).
\newblock


\bibitem[\protect\citeauthoryear{Ji, He, Xu, Liu, and Zhao}{Ji
  et~al\mbox{.}}{2015}]%
        {Ji:1}
\bibfield{author}{\bibinfo{person}{Guoliang Ji}, \bibinfo{person}{Shizhu He},
  \bibinfo{person}{Liheng Xu}, \bibinfo{person}{Kang Liu}, {and}
  \bibinfo{person}{Jun Zhao}.} \bibinfo{year}{2015}\natexlab{}.
\newblock \showarticletitle{Knowledge graph embedding via dynamic mapping
  matrix}.
\newblock \bibinfo{journal}{{\em In Proceedings of the 53rd Annual Meeting of
  the Association for Computational Linguistics and the 7th International Joint
  Conference on Natural Language Processing\/}} (\bibinfo{year}{2015}),
  \bibinfo{pages}{687--696}.
\newblock


\bibitem[\protect\citeauthoryear{Lin, Liu, Luan, Sun, Rao, and Liu}{Lin
  et~al\mbox{.}}{2015a}]%
        {Lin:2015b}
\bibfield{author}{\bibinfo{person}{Yankai Lin}, \bibinfo{person}{Zhiyuan Liu},
  \bibinfo{person}{Huanbo Luan}, \bibinfo{person}{Maosong Sun},
  \bibinfo{person}{Siwei Rao}, {and} \bibinfo{person}{Song Liu}.}
  \bibinfo{year}{2015}\natexlab{a}.
\newblock \showarticletitle{Modeling Relation Paths for Representation Learning
  of Knowledge Bases}.
\newblock \bibinfo{journal}{{\em In Proceedings of the 2015 Conference on
  Empirical Methods in Natural Language Processing\/}} (\bibinfo{year}{2015}),
  \bibinfo{pages}{705--714}.
\newblock


\bibitem[\protect\citeauthoryear{Lin, Liu, Sun, Liu, and Zhu}{Lin
  et~al\mbox{.}}{2015b}]%
        {Lin:2015a}
\bibfield{author}{\bibinfo{person}{Yankai Lin}, \bibinfo{person}{Zhiyuan Liu},
  \bibinfo{person}{Maosong Sun}, \bibinfo{person}{Yang Liu}, {and}
  \bibinfo{person}{Xuan Zhu}.} \bibinfo{year}{2015}\natexlab{b}.
\newblock \showarticletitle{Learning entity and relation embeddings for
  knowledge graph completion}.
\newblock \bibinfo{journal}{{\em In Proceedings of the Twenty-Ninth AAAI
  Conference on Artificial Intelligence\/}} (\bibinfo{year}{2015}),
  \bibinfo{pages}{2181--2187}.
\newblock


\bibitem[\protect\citeauthoryear{Luo, Wang, Wang, and Guo}{Luo
  et~al\mbox{.}}{2015}]%
        {Luo:2015}
\bibfield{author}{\bibinfo{person}{Yuanfei Luo}, \bibinfo{person}{Quan Wang},
  \bibinfo{person}{Bin Wang}, {and} \bibinfo{person}{Li Guo}.}
  \bibinfo{year}{2015}\natexlab{}.
\newblock \showarticletitle{Context-dependent knowledge graph embedding}.
\newblock \bibinfo{journal}{{\em In Proceedings of the 2015 Conference on
  Empirical Methods in Natural Language Processing\/}} (\bibinfo{year}{2015}),
  \bibinfo{pages}{1656--1661}.
\newblock


\bibitem[\protect\citeauthoryear{Miller}{Miller}{1995}]%
        {Miller:1}
\bibfield{author}{\bibinfo{person}{George~A. Miller}.}
  \bibinfo{year}{1995}\natexlab{}.
\newblock \showarticletitle{WordNet: a lexical database for English}.
\newblock \bibinfo{journal}{{\it Commun. ACM}} \bibinfo{volume}{38},
  \bibinfo{number}{11} (\bibinfo{year}{1995}), \bibinfo{pages}{39--41}.
\newblock


\bibitem[\protect\citeauthoryear{Nickel, Rosasco, and Poggio}{Nickel
  et~al\mbox{.}}{2015}]%
        {Nickel:2015}
\bibfield{author}{\bibinfo{person}{Maximilian Nickel}, \bibinfo{person}{Lorenzo
  Rosasco}, {and} \bibinfo{person}{Tomaso Poggio}.}
  \bibinfo{year}{2015}\natexlab{}.
\newblock \showarticletitle{Holographic embeddings of knowledge graphs}.
\newblock \bibinfo{journal}{{\em Computer Science\/}} (\bibinfo{year}{2015}).
\newblock


\bibitem[\protect\citeauthoryear{Nickel, Tresp, and Kriegel}{Nickel
  et~al\mbox{.}}{2011}]%
        {Nickel:2011}
\bibfield{author}{\bibinfo{person}{Maximilian Nickel}, \bibinfo{person}{Volker
  Tresp}, {and} \bibinfo{person}{Hans-Peter Kriegel}.}
  \bibinfo{year}{2011}\natexlab{}.
\newblock \showarticletitle{A three-way model for collective learning on
  multi-relational data}.
\newblock \bibinfo{journal}{{\em In International Conference on Machine
  Learning\/}} (\bibinfo{year}{2011}), \bibinfo{pages}{809--816}.
\newblock


\bibitem[\protect\citeauthoryear{Niu and Li}{Niu and Li}{2017}]%
        {Niu:2017}
\bibfield{author}{\bibinfo{person}{Xiao-Fan Niu} {and} \bibinfo{person}{Wu-Jun
  Li}.} \bibinfo{year}{2017}\natexlab{}.
\newblock \showarticletitle{ParaGraphE: A Library for Parallel Knowledge Graph
  Embedding}.
\newblock \bibinfo{journal}{{\em arXiv\/}} (\bibinfo{year}{2017}).
\newblock


\bibitem[\protect\citeauthoryear{Pujara, Miao, Getoor, and Cohen}{Pujara
  et~al\mbox{.}}{2013}]%
        {Pujara:2013}
\bibfield{author}{\bibinfo{person}{Jay Pujara}, \bibinfo{person}{Hui Miao},
  \bibinfo{person}{Lise Getoor}, {and} \bibinfo{person}{William Cohen}.}
  \bibinfo{year}{2013}\natexlab{}.
\newblock \showarticletitle{Knowledge graph identification}.
\newblock \bibinfo{journal}{{\em In Proceedings of the 12th International
  Semantic Web Conference\/}} (\bibinfo{year}{2013}),
  \bibinfo{pages}{542--557}.
\newblock


\bibitem[\protect\citeauthoryear{Riedel, Yao, Mccallum, and Marlin}{Riedel
  et~al\mbox{.}}{2013}]%
        {Riedel:2013}
\bibfield{author}{\bibinfo{person}{Sebastian Riedel}, \bibinfo{person}{Limin
  Yao}, \bibinfo{person}{Andrew Mccallum}, {and} \bibinfo{person}{Benjamin~M
  Marlin}.} \bibinfo{year}{2013}\natexlab{}.
\newblock \showarticletitle{Relation extraction with matrix factorization and
  universal schemas}.
\newblock \bibinfo{journal}{{\em In Proceedings of NAACL-HLT\/}}
  (\bibinfo{year}{2013}), \bibinfo{pages}{74--84}.
\newblock


\bibitem[\protect\citeauthoryear{Rockt\"aschel, Singh, and
  Riedel}{Rockt\"aschel et~al\mbox{.}}{2015}]%
        {Rocktaschel:2015}
\bibfield{author}{\bibinfo{person}{Tim Rockt\"aschel}, \bibinfo{person}{Sameer
  Singh}, {and} \bibinfo{person}{Sebastian Riedel}.}
  \bibinfo{year}{2015}\natexlab{}.
\newblock \showarticletitle{Injecting Logical Background Knowledge into
  Embeddings for Relation Extraction}.
\newblock \bibinfo{journal}{{\em In Annual Conference of the North American
  Chapter of the Association for Computational Linguistics\/}}
  (\bibinfo{year}{2015}).
\newblock


\bibitem[\protect\citeauthoryear{Socher, Chen, Manning, and Ng}{Socher
  et~al\mbox{.}}{2013}]%
        {Socher:2013}
\bibfield{author}{\bibinfo{person}{Richard Socher}, \bibinfo{person}{Danqi
  Chen}, \bibinfo{person}{Christopher~D Manning}, {and} \bibinfo{person}{Andrew
  Ng}.} \bibinfo{year}{2013}\natexlab{}.
\newblock \showarticletitle{Reasoning with neural tensor networks for knowledge
  base completion}.
\newblock \bibinfo{journal}{{\em In Advances in Neural Information Processing
  Systems\/}} (\bibinfo{year}{2013}), \bibinfo{pages}{926--934}.
\newblock


\bibitem[\protect\citeauthoryear{Suchanek, Kasneci, and Weikum}{Suchanek
  et~al\mbox{.}}{2007}]%
        {Suchanek:1}
\bibfield{author}{\bibinfo{person}{Fabian~M. Suchanek},
  \bibinfo{person}{Gjergji Kasneci}, {and} \bibinfo{person}{Gerhard Weikum}.}
  \bibinfo{year}{2007}\natexlab{}.
\newblock \showarticletitle{Yago: a core of semantic knowledge}.
\newblock \bibinfo{journal}{{\em In Proceedings of the 16th international
  conference on World Wide Web\/}} (\bibinfo{year}{2007}),
  \bibinfo{pages}{697--706}.
\newblock


\bibitem[\protect\citeauthoryear{Sutskever, Salakhutdinov, and
  Tenenbaum.}{Sutskever et~al\mbox{.}}{2009}]%
        {Sutskever:2009}
\bibfield{author}{\bibinfo{person}{Ilya Sutskever},
  \bibinfo{person}{Ruslan~Ruslan Salakhutdinov}, {and}
  \bibinfo{person}{Joshua~B. Tenenbaum.}} \bibinfo{year}{2009}\natexlab{}.
\newblock \showarticletitle{Modelling relational data using bayesian clustered
  tensor factorization}.
\newblock \bibinfo{journal}{{\em In Proceedings of the 23rd Annual Conference
  on Neural Information Processing Systems\/}} (\bibinfo{year}{2009}),
  \bibinfo{pages}{1821--1828}.
\newblock


\bibitem[\protect\citeauthoryear{Wang, Wang, and Guo}{Wang
  et~al\mbox{.}}{2015}]%
        {Wang:2015}
\bibfield{author}{\bibinfo{person}{Quan Wang}, \bibinfo{person}{Bin Wang},
  {and} \bibinfo{person}{Li Guo}.} \bibinfo{year}{2015}\natexlab{}.
\newblock \showarticletitle{Knowledge base completion using embeddings and
  rules}.
\newblock \bibinfo{journal}{{\em In Proceedings of the 24th International
  Conference on Artificial Intelligence\/}} (\bibinfo{year}{2015}),
  \bibinfo{pages}{1859--1865}.
\newblock


\bibitem[\protect\citeauthoryear{Wang, Zhang, Feng, and Chen}{Wang
  et~al\mbox{.}}{2014a}]%
        {Wang:2014b}
\bibfield{author}{\bibinfo{person}{Zhen Wang}, \bibinfo{person}{Jianwen Zhang},
  \bibinfo{person}{Jianlin Feng}, {and} \bibinfo{person}{Zheng Chen}.}
  \bibinfo{year}{2014}\natexlab{a}.
\newblock \showarticletitle{Knowledge graph and text jointly embedding}.
\newblock \bibinfo{journal}{{\em In Proceedings of the 2014 Conference on
  Empirical Methods in Natural Language Processing\/}} (\bibinfo{year}{2014}),
  \bibinfo{pages}{1591--1601}.
\newblock


\bibitem[\protect\citeauthoryear{Wang, Zhang, Feng, and Chen}{Wang
  et~al\mbox{.}}{2014b}]%
        {Wang:2014a}
\bibfield{author}{\bibinfo{person}{Zhen Wang}, \bibinfo{person}{Jianwen Zhang},
  \bibinfo{person}{Jianlin Feng}, {and} \bibinfo{person}{Zheng Chen}.}
  \bibinfo{year}{2014}\natexlab{b}.
\newblock \showarticletitle{Knowledge graph embedding by translating on
  hyperplanes}.
\newblock \bibinfo{journal}{{\em In Proceedings of the Twenty-Eighth AAAI
  Conference on Artificial Intelligence\/}} (\bibinfo{year}{2014}),
  \bibinfo{pages}{1112--1119}.
\newblock


\bibitem[\protect\citeauthoryear{Wei, Zhao, Liu, Qi, Sun, and Tian}{Wei
  et~al\mbox{.}}{2015}]%
        {Wei:2015}
\bibfield{author}{\bibinfo{person}{Zhuoyu Wei}, \bibinfo{person}{Jun Zhao},
  \bibinfo{person}{Kang Liu}, \bibinfo{person}{Zhenyu Qi},
  \bibinfo{person}{Zhengya Sun}, {and} \bibinfo{person}{Guanhua Tian}.}
  \bibinfo{year}{2015}\natexlab{}.
\newblock \showarticletitle{Large-scale Knowledge Base Completion: Inferring
  via Grounding Network Sampling over Selected Instances}.
\newblock \bibinfo{journal}{{\em In Proceedings of the 24th ACM International
  on Conference on Information and Knowledge Management\/}}
  (\bibinfo{year}{2015}), \bibinfo{pages}{1331--1340}.
\newblock


\bibitem[\protect\citeauthoryear{Weston, Bordes, Yakhnenko, and Usunier}{Weston
  et~al\mbox{.}}{2013}]%
        {Weston:2013}
\bibfield{author}{\bibinfo{person}{Jason Weston}, \bibinfo{person}{Antoine
  Bordes}, \bibinfo{person}{Oksana Yakhnenko}, {and} \bibinfo{person}{Nicolas
  Usunier}.} \bibinfo{year}{2013}\natexlab{}.
\newblock \showarticletitle{Connecting Language and Knowledge Bases with
  Embedding Models for Relation Extraction}.
\newblock \bibinfo{journal}{{\em In Proceedings of the 2013 Conference on
  Empirical Methods in Natural Language Processing\/}} (\bibinfo{year}{2013}),
  \bibinfo{pages}{1366--1371}.
\newblock


\bibitem[\protect\citeauthoryear{Wu, Li, Wang, and Zhu}{Wu
  et~al\mbox{.}}{2012}]%
        {Wu:1}
\bibfield{author}{\bibinfo{person}{Wentao Wu}, \bibinfo{person}{Hongsong Li},
  \bibinfo{person}{Haixun Wang}, {and} \bibinfo{person}{Kenny~Q. Zhu}.}
  \bibinfo{year}{2012}\natexlab{}.
\newblock \showarticletitle{Probase: A probabilistic taxonomy for text
  understanding.}
\newblock \bibinfo{journal}{{\em In Proceedings of the 2012 ACM SIGMOD
  International Conference on Management of Data\/}} (\bibinfo{year}{2012}),
  \bibinfo{pages}{481--492}.
\newblock


\bibitem[\protect\citeauthoryear{Xu, Tresp, Yu, and peter Kriegel}{Xu
  et~al\mbox{.}}{2006}]%
        {Xu:2006}
\bibfield{author}{\bibinfo{person}{Zhao Xu}, \bibinfo{person}{Volker Tresp},
  \bibinfo{person}{Kai Yu}, {and} \bibinfo{person}{Hans peter Kriegel}.}
  \bibinfo{year}{2006}\natexlab{}.
\newblock \showarticletitle{Infinite hidden relational models.}
\newblock \bibinfo{journal}{{\em In Proceedings of the 22nd Conference on
  Uncertainty in Artificial Intelligence\/}} (\bibinfo{year}{2006}),
  \bibinfo{pages}{544--551}.
\newblock


\bibitem[\protect\citeauthoryear{Zhong, Zhang, Wang, Wan, and Chen}{Zhong
  et~al\mbox{.}}{2015}]%
        {Zhong:2015}
\bibfield{author}{\bibinfo{person}{Huaping Zhong}, \bibinfo{person}{Jianwen
  Zhang}, \bibinfo{person}{Zhen Wang}, \bibinfo{person}{Hai Wan}, {and}
  \bibinfo{person}{Zheng Chen}.} \bibinfo{year}{2015}\natexlab{}.
\newblock \showarticletitle{Aligning knowledge and text embeddings by entity
  descriptions.}
\newblock \bibinfo{journal}{{\em In Proceedings of the 2015 Conference on
  Empirical Methods in Natural Language Processing\/}} (\bibinfo{year}{2015}),
  \bibinfo{pages}{267--272}.
\newblock


\end{thebibliography}
